\documentclass{article}

\usepackage{microtype}
\usepackage{graphicx}
\usepackage{booktabs} 
\usepackage{amsmath}
\usepackage{amsfonts}
\usepackage{mathtools}
\usepackage{autobreak}
\usepackage{breqn}
\usepackage{xcolor, soul}
\usepackage{makecell}
\usepackage{algorithmic}
\usepackage{siunitx}
\usepackage{caption}
\usepackage{subcaption}
\usepackage{multirow}

\usepackage{pgfplots}
\pgfplotsset{width=10cm,compat=1.9}
\usepgfplotslibrary{external}
\usetikzlibrary{fillbetween}
\usepgfplotslibrary{dateplot}

\newsavebox\TBox
\def\textoverline#1{\savebox\TBox{#1}%
  \makebox[0pt][l]{#1}\rule[1.1\ht\TBox]{\wd\TBox}{0.4pt}}

\definecolor{bluebox}{RGB}{218,232,252}
\definecolor{greenbox}{RGB}{213,232,212}
\definecolor{bluebb}{RGB}{138,166,206}
\definecolor{greenbb}{RGB}{146,188,121}

\newcommand{\argmax}{\mathop{\rm arg~max}\limits}

\renewcommand{\algorithmiccomment}[1]{\bgroup\hfill~#1\egroup}

\usepackage{hyperref}

\usepackage[nameinlink,capitalize]{cleveref}

\usepackage[accepted]{icml2021}

\icmltitlerunning{A Unified Generative Adversarial Network Training}

\begin{document}

\twocolumn[

\icmltitle{A Unified Generative Adversarial Network Training\\ via Self-Labeling and Self-Attention}

\icmlsetsymbol{equal}{*}

\begin{icmlauthorlist}
\icmlauthor{Tomoki Watanabe}{toshiba}
\icmlauthor{Paolo Favaro}{unibe}
\end{icmlauthorlist}

\icmlaffiliation{unibe}{University of Bern, Bern, Switzerland}
\icmlaffiliation{toshiba}{Toshiba Corporation, Kawasaki, Japan}

\icmlcorrespondingauthor{Tomoki Watanabe}{tomoki8.watanabe@toshiba.co.jp}
\icmlcorrespondingauthor{Paolo Favaro}{paolo.favaro@inf.unibe.ch}

\icmlkeywords{Generative Adversarial Network, Machine Learning, ICML}

\vskip 0.3in
]

\printAffiliationsAndNotice{This work has been done while the first author was visiting the University of Bern.}  

\begin{abstract}
We propose a novel GAN training scheme that can handle any level of labeling in a unified manner.
Our scheme introduces a form of artificial labeling that can incorporate manually defined labels, when available, and induce an alignment between them.
To define the artificial labels, we exploit the assumption that neural network generators can be trained more easily to map nearby latent vectors to data with semantic similarities, than across separate categories.
We use generated data samples and their corresponding artificial conditioning labels to train a classifier. 
The classifier is then used to self-label real data. To boost the accuracy of the self-labeling, we also use the exponential moving average of the classifier.
However, because the classifier might still make  mistakes, especially at the beginning of the training, we also refine the labels through self-attention, by using the labeling of real data samples only when the classifier outputs a high classification probability score.
We evaluate our approach on CIFAR-10, STL-10 and SVHN, and show that both self-labeling and self-attention consistently 
improve the quality of generated data. More surprisingly, we find that the proposed scheme can even outperform class-conditional GANs. 
\end{abstract}

\section{Introduction}

\begin{figure*}[t]
\centering
\centerline{\includegraphics[scale=0.94]{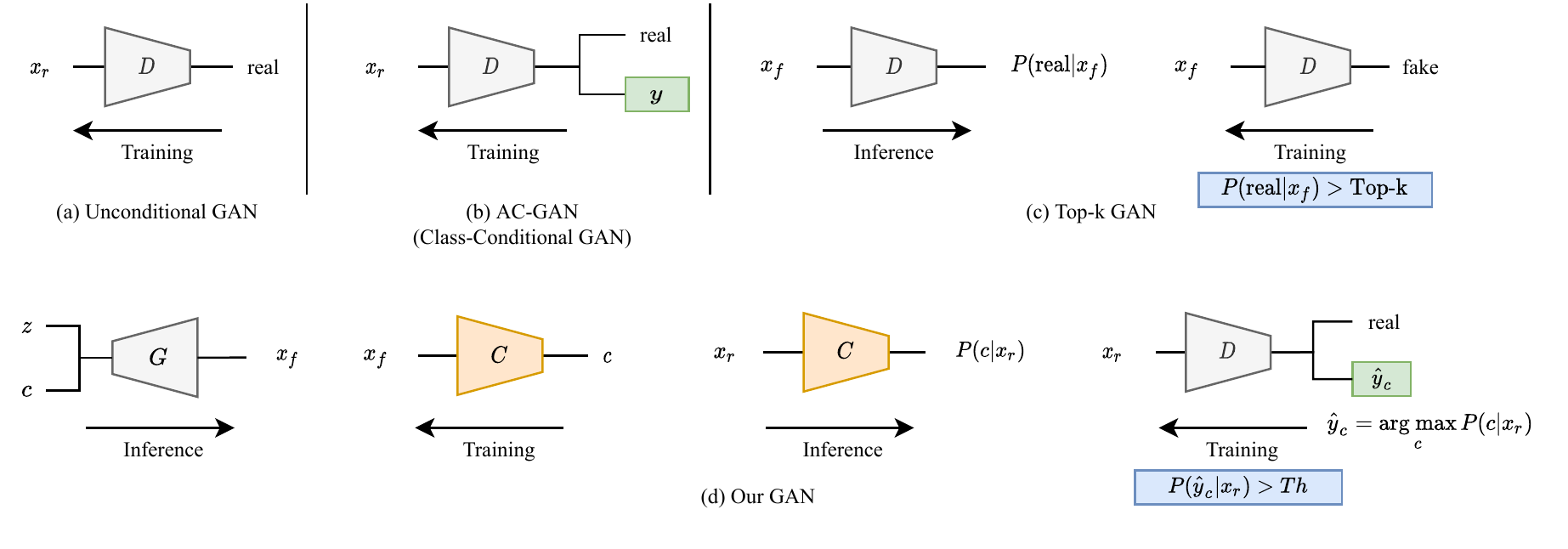}}
\vskip -0.2in
\caption{{\bf Comparison of GAN variants and our GAN training scheme}: (a) Unconditional GAN discriminates real samples $x_r$ and fake samples $x_f$; (b) AC-GAN~\cite{ac_gan} learns to predict {\it supervised} labels $y$; (C) Top-k GAN~\cite{ktop_gan} updates the discriminator $D$ with fake samples where the discriminator $D$ outputs high confidence, \emph{i.e.}, $P(\text{real}|x_f)>\text{Top-k}$; (d) Our proposed GAN scheme learns to predict {\it artificial} labels $\hat{y}_c$ defined by the teacher classifier $C$ and updates the discriminator $D$ on {\it real} samples, where the teacher classifier $C$ outputs high confidence, \emph{i.e.}, $P(c|x_r)\ge Th$. We train the teacher classifier only on fake samples $x_f$ and conditional labels $c$ of the generator $G$.}
\label{fig:compare}
\vskip -0.1in
\end{figure*}

Generative Adversarial Networks (GAN)~\cite{gan,biggan,style_gan} provide an attractive approach to constructing generative models that output samples of a target distribution. 
In their most basic form, these models consist of two neural networks, a generator and a discriminator. The first network is trained to generate samples from some latent representation (typically a sample from a Gaussian distribution), while the second network is trained to distinguish real samples $x_r$ from the generated samples $x_f$ (see network $D$ in \autoref{fig:compare}~(a)).
The most effective GANs seem to benefit greatly from class conditioning. 
The class information is provided as input to the generator and either injected into the discriminator as an input~\cite{mizraOsindero2018}
or through intermediate layers~\cite{reedLee2016}
or via a projection~\cite{proj_gan} or an auxiliary loss~\cite{ac_gan, mhinge_gan} (see \emph{e.g.}, \autoref{fig:compare}~(b)).
The family of these generators is generically called \emph{conditional GANs} (cGAN).

In particular, in AC-GAN \cite{ac_gan} (see \autoref{fig:compare}~(b)) one uses a discriminator with two outputs, one for the classification of input images into real or fake and the other for classification into multiple categories (see the label $y$ in the \fcolorbox{greenbb}{greenbox}{green box} in \autoref{fig:compare}~(b)). We hypothesize that providing the class information helps the training of the generator, because the neural network architecture of the generator tends to map similar latent vectors to data samples that are semantically related. Thus, we expect a gradient-based training to converge more easily to a good set of network parameters with class conditioning than without it. This suggests that one might be able to train a generator in a conditional manner even when manually defined labels are not available.
Based on this assumption, we train a generator conditioned on an artificial set of labels and simultaneously learn its inverse mapping through a classifier, which we call \emph{teacher}.
The teacher is trained only on synthetic data (networks $G$-inference and $C$-training in \autoref{fig:compare}~(d)), and then it is applied to real data samples to obtain their corresponding artificial labels (network $C$-inference in \autoref{fig:compare}~(d)), a process that we call \emph{self-labeling}.
Because the accuracy of a trained classifier is limited, using all the artificially labeled real data would not help the generator especially at the beginning of the training, when all predicted labels may be highly inaccurate. Hence, we introduce a way to select data, where the label consistency is high. To do so we introduce a \emph{self-attention} mechanism that is based on selecting samples $x_r$ whose estimated label probability is above a given threshold (\fcolorbox{bluebb}{bluebox}{blue box} in \autoref{fig:compare}~(d)). The selected samples and their corresponding synthetic labels $\hat{y}_c$ are then used to train the discriminator (network $D$ and \fcolorbox{greenbb}{greenbox}{green box} in \autoref{fig:compare}~(d)). 
This idea is similar to the one exploited by Top-k GAN~\cite{ktop_gan}, which uses the output of the discriminator as the confidence for the labeling of fake images (see \autoref{fig:compare}~(c)), while we apply it instead to real images.
Moreover, we use the EMA~(exponential moving average) of the teacher during inference to further improve its labeling accuracy. This technique has been used in semi-supervised and unsupervised learning methods to improve the classification/clustering accuracy~\cite{mean_teacher, fixmatch, moco}.

So far, we have described a GAN training method that exploits the same benefits that conditional GANs enjoy, but without using manually labeled data. When data is partially or fully labeled, it is desirable to take advantage of the available information. Our scheme can seamlessly integrate such available labels and also indirectly transfer their categorical information to the artificial labels. This is possible because our artificial labels are defined relative to the generator and the generator can adapt to a new reference during training.

We evaluate our method on CIFAR-10~\cite{cifar10}, STL-10~\cite{stl10}, and SVHN~\cite{svhn} datasets using the BigGAN model~\cite{biggan} and show that our method improves the quality of the generated images in terms of the FID (Fr\'echet Inception Distance) score \cite{tf_fid}.
Our method achieves better FID scores than the state-of-the-art GAN and even that of fully supervised cGAN methods on the CIFAR-10 dataset.
Our contributions can be summarized as follows:\\
\noindent\textbf{1)} A unified GAN training that can handle any level of labeling in a unified manner by using: 
{\bf Self-labeling}: a method to automatically assign labels to real data samples, and 
{\bf Self-attention}: a method to select real data samples with highly consistent synthetic labels;\\
\noindent\textbf{2)} Consistent improvement in the FID scores across several datasets (evaluation on CIFAR-10, STL-10, and SVHN);\\
\noindent\textbf{3)} The ability to outperform class-conditional GANs (fully labeled dataset). 

\section{Prior Work}

\begin{figure}[t]
\centering
\centerline{\includegraphics[scale=0.96]{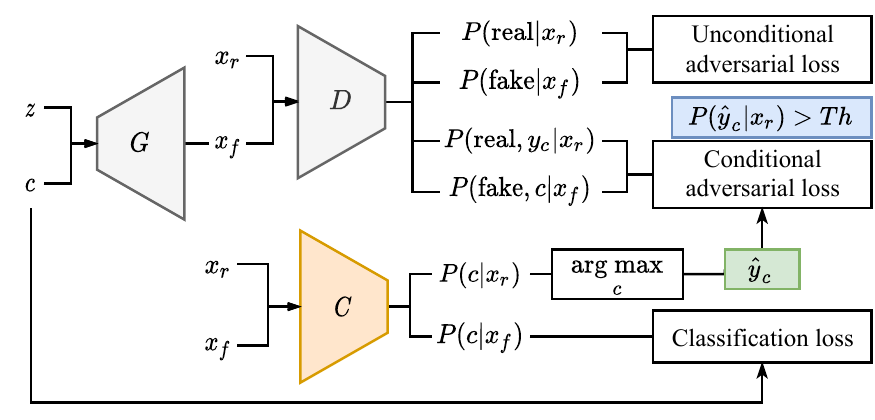}}
\caption{{\bf Overall architecture of our proposed unified GAN training.} The important components of our scheme are: 1) The conditional training of the generator $G$, which is based on artificial labels $c$; 2) A classifier $C$, which is trained only on synthetic data samples $x_f$ and their corresponding artificial labels $c$; 3) A discriminator $D$ that is trained on artificial labels $c$ for fake data $x_f$, real labels $y$ for real data $x_r$ when available, and generated labels $\hat{y}_c$ for real data $x_r$ that do not have manually defined labels.}
\label{fig:network}
\vskip -0.1in
\end{figure}

{\bf Generative Adversarial Networks.} The unconditional GAN~\cite{biggan} consists of two networks, a generator $G$ and a discriminator $D$. The generator outputs a fake image from a (vector) noise instance and the discriminator distinguishes input images as real or fake. The discriminator and the generator are trained by minimizing the following adversarial loss terms
\begin{align}
\nonumber
        L_D^U = 
        &-E_{x_r}[\log P(\text{real}|x_r)]\\
            \label{eq:LDU}
        &- E_z[\log P(\text{fake}|G(z))],\\
    \label{eq:LGU}
        L_G^U = 
        &- E_z[\log P(\text{real}|G(z))],
\end{align}
where $x_r$ is a random variable representing real images (and is used also to indicate samples from the distribution of real images) and $z$ is a random variable with a fixed distribution, typically the Normal distribution ${\cal N}(0,I_d)$ (and is also used to indicate instances from that distribution). $P(\text{class}|\text{input})$ indicates the probability that \texttt{input} belongs to the class \texttt{class}.
Class-conditional GANs (\emph{e.g.}, AC-GAN~\cite{ac_gan}) are instead trained by minimizing the following adversarial loss terms
\begin{align}
\nonumber
        L_D^Y = 
        &-E_{x_r,y}[\log P(\text{real}, y|x_r)]\\
            \label{eq:LDY}
        &-E_{z,c}[\log P(\text{fake}, c|G(z,c))],\\
    \label{eq:LGC}
        L_G^C = 
        &-E_{z,c}[\log P(\text{real}, c|G(z,c))],
\end{align}
where $y$ and $c$ are supervised labels and artificial conditional labels respectively.
The discriminator of cGAN does not only classify images into real/false, but also estimates the labels of the input images.
The training requires supervised labels for all real images.
An alternative approach to using real labels is proposed in self-conditional GANs, which are cGANs trained with unlabeled samples, where artificial labels are defined through some heuristic tasks.
An example of this approach is to use the orientation of rotated images as a label for conditioning~\cite{ssgan_rotation, ssgan_minimax}. In comparison to these methods, we obtain our labels implicitly from the generator.

{\bf Semi-Supervised Learning.} Semi-supervised learning~(SSL) is a branch of machine learning where the training data is a mix of labeled samples (typically, a small amount) and of unlabeled samples (typically, in larger number compared to the labeled samples) \cite{SSLsurvey2020}. 
Recent methods that work in the SSL regime are RemixMatch~\cite{remixmatch} and FixMatch~\cite{fixmatch}.

In particular, of interest to us is FixMatch~\cite{fixmatch}, which trains a classifier $C$ by minimizing two cross-entropy loss terms: a supervised loss on labeled samples $x^L$ and an unsupervised loss on unlabeled samples $x^U$ as 
\begin{align}
    \begin{autobreak}
        L_{label} =
        H[y, C(A(x^L))]
        +H[\hat{y}, C(A(x^U))],
    \end{autobreak}
\end{align}
where $y$ and $\hat{y}$ are supervised labels and artificial labels respectively, and $A$ is an image augmentation function.
As in our approach, the artificial labels are assigned by a classifier with the parameters $\Bar{\theta}_C$ of the running average model \cite{mean_teacher} via
\begin{align}
    \begin{autobreak}
        \hat{y} = 
        \argmax_i C_i(\alpha(x^U); \Bar{\theta}_C).
    \end{autobreak}
\end{align}
where $\alpha$ is a \emph{weak} image augmentation function, \emph{i.e.}, with image transformations close to the identity. 

Other important recent SSL methods for GAN training are the work of~\cite{s3gan,slcgan}.
As in our approach, they train a network for classification/clustering, which is then used to provide labels for conditioning.
However, while they train the classifier on real samples, we train it only on fake samples, and, to the best of our knowledge, are the first ones to do so.

\section{A Unified GAN Training}

Our unified GAN training uses a cGAN as backbone, where the discriminator classifies the input into real/fake and image categories. cGANs require semantic labels for training. While the labels of generated data are implicitly defined, the labels of real data are either provided through manual labeling or through our unsupervised self-labeling and self-attention procedures. 

\subsection{Self-Labeling and Self-Attention}
\label{sec:teacher}
Our objective is to assign artificial labels to unlabeled real images that are used in the conditional adversarial loss. 
To this purpose, we train a classifier $C$, which we call \emph{teacher}, on fake images $x_f = G(z, c)$, where the class-correspondence is known. 
We train the teacher with the cross-entropy loss
\begin{align}
    \label{eq:LC}
    \begin{autobreak}
        L_C = 
        H[c, C(A(x_f))].
    \end{autobreak}
\end{align}
where $H$ denotes the entropy, the fake image is obtained via $x_f = G(z,c)$ with $z\sim {\cal N}(0,I_d)$, and $c$ is a random variable (with a discrete Uniform distribution) and also denotes its instance.
Since the teacher may not be a perfect inverse of $G$ with respect to the conditional label $c$, we introduce two methods to ensure a high classification accuracy.

First, we use the EMA parameters of the teacher to compute the artificial labels $\hat{y}_c$ of real images, \emph{i.e.}, we compute
\begin{align}
    \label{eq:yc}
    \begin{autobreak}
        \hat{y}_c = 
        \argmax_i C_i(\alpha(x_r); \bar{\theta}_C).
    \end{autobreak}
\end{align}
Second, because the artificial labels $\hat{y}_c$ are inaccurate especially during the early epochs of the training, we introduce a selection mechanism called self-attention. 
We first define the \emph{reliability} of the artificial labels via the softmax of the classifier output
\begin{align}
    \label{eq:pc}
    p_c = \frac{\exp(C_{\hat{y}_c}(\alpha(x_r); \bar{\theta}_C))}{\sum_{i=1}^K \exp(C_i(\alpha(x_r); \bar{\theta}_C))},
\end{align}
where $K$ is the number of the artificial classes. As we show in the experiments, the reliability yields a high value with real images that are distinctively similar to generated fake images, and when these fake images are well separated into different clusters. Then, self-attention selects real images $x_r$ such that $p_c \ge Th$, where the threshold $Th\in[0,1]$.

\subsection{Training with Artificial (and Real) Labels}
\label{sec:cgan_al}

The conditional adversarial loss for conventional cGANs uses the supervised class labels $y$ and artificial labels $c$ for real images and fake images respectively as shown in \cref{eq:LDY}.
With real images without supervised class labels $y$, we use instead the artificial labels $\hat{y}_c$.

The discriminator has 2 heads, one for the unconditional fake/real adversarial loss and another for the conditional adversarial loss.
The losses for the discriminator $L_D$ and the generator $L_G$ are simply the sum of the corresponding conditional and unconditional losses
\begin{align}
    \label{eq:LD}
        L_D = 
        L_D^U + L_D^C ,\quad
        L_G = 
        L_G^U + L_G^C .
\end{align}
The loss functions $L_D^U$, $L_G^U$, and $L_G^C$ are shown in \cref{eq:LDU,eq:LGU,eq:LGC}. The loss function~$L_D^C$ instead is defined so that it can be applied to a dataset with any degree of labeling (from $0\%$ to $100\%$) as 
\begin{align}
    \label{eq:LDC}
    \begin{autobreak}
        L_D^C = 
        -E_{\{x_r,y|\text{with label}\}}[\log P(\text{real}, y|x_r)] 
        -E_{\{x_r, \hat{y}_c|\text{no label } \wedge~p_c\ge Th\}}[\log P(\text{real}, \hat{y}_c|x_r)] 
        - E_{x_f,c}[\log P(\text{fake}, c|x_f)] .
    \end{autobreak}
\end{align}
The loss function uses artificial labels $\hat{y}_c$ obtained from the teacher as shown in \cref{eq:yc}. As explained in \autoref{sec:teacher}, we calculate the loss only on images where the reliability $p_c$ is higher than a threshold $Th$, because unreliable labels have an adverse effect on the training of the discriminator.
We update the teacher, the discriminator, and the generator simultaneously via \cref{eq:LC,eq:pc,eq:LD}. 
We can train cGAN on unlabeled dataset, because these loss terms are well-defined even in the absence of real labels.

We show the network architecture of our method in \autoref{fig:network}. The components were already introduced in  \autoref{fig:compare}~(d).

\begin{algorithm}[tb]
  \caption{Unified GAN Training}
  \label{alg:ours}
\begin{algorithmic}
  \STATE {\bfseries Input:} Parameters of the generator $\theta_G$, the discriminator $\theta_D$, the teacher $\theta_C$, the number of artificial classes $K$, and the threshold $Th$
  \FOR{the number of training iterations}
  \STATE Sample batch $z \sim p(z)$, $c \sim p(c)$, $x_r \sim p_{real}(x_r)$
  \STATE {\bf Step1. Update teacher:}
    \COMMENT{\autoref{sec:teacher}}
  \STATE $L_C \leftarrow$ SoftmaxCrossEntropy$(c, C(A(x_f))$
    \COMMENT{\cref{eq:LC}}
  \STATE $\theta_C \leftarrow$ MomentumOptimizer($L_C$)
  \STATE $\bar{\theta}_C \leftarrow$ ExponentialMovingAverage($\theta_C$)
  \STATE $\hat{y}_c \leftarrow \argmax{}_i C_i(\alpha(x_r); \Bar{\theta}_C)$ 
    \COMMENT{(self-labeling) \cref{eq:yc}}
  \STATE $p_c \leftarrow$ Softmax$(C_{\hat{y}_c}(\alpha(x_r); \Bar{\theta}_C))$
    \COMMENT{\cref{eq:pc}}
  \STATE {\bf Step 2. Update cGAN:}
    \COMMENT{\autoref{sec:cgan_al}}
  \STATE $x_f \leftarrow G(c, z)$
  \STATE $L_D^U \leftarrow$ Hinge$(D^U(x_r)) +$ Hinge$(-D^U(x_f))$ 
    \COMMENT{\cref{eq:LDU}}
  \STATE $S \leftarrow$ {\bf if} $p_c \ge Th$ {\bf then} $1$ {\bf else} $0$
    \COMMENT{(self-attention)}
  \STATE $L_D^C \leftarrow S\cdot$ MultiClassHinge$(\hat{y}_c, D^C(x_r))$
  \STATE \quad\qquad $+$ MultiClassHinge$(c+K, D^C(x_f))$ 
    \COMMENT{\cref{eq:LDC}}
  \STATE $\theta_D \leftarrow$ AdamOptimizer($L_D^U + L_D^C$)
  \STATE $L_G^U \leftarrow$ Hinge$(D^U(G(c,z)))$ 
    \COMMENT{\cref{eq:LGU}}
  \STATE $L_G^C \leftarrow$ MultiClassHinge$(c, D^C(G(c,z))$ 
    \COMMENT{\cref{eq:LGC}}
  \STATE $\theta_G \leftarrow$ AdamOptimizer($L_G^U + L_G^C$)
  \ENDFOR
\end{algorithmic}
\end{algorithm}

\begin{figure*}[t]
\centering
\subfloat[Real images]{
\hspace{-1.8mm}
\includegraphics[width=.49\linewidth]{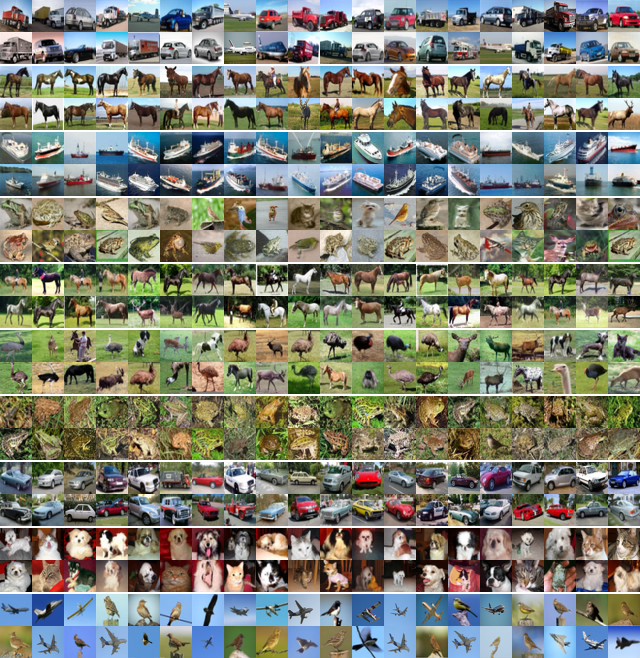}
\label{fig:reals_unlabeled}
}
\subfloat[Fake images]{
\includegraphics[width=.49\linewidth]{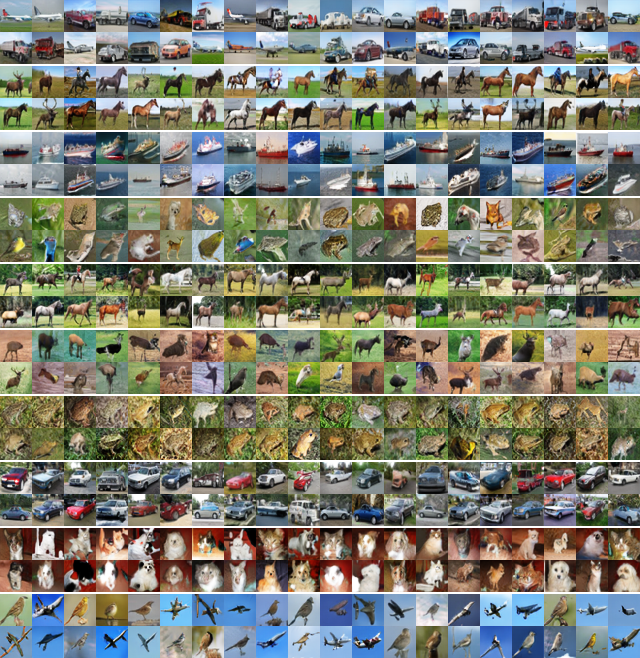}
\label{fig:fakes_unlabeled}
}
\caption{{\bf (a)~Real images and (b)~Fake images on the unlabeled CIFAR-10.} From the top to the bottom, every pair of image rows corresponds to the same artificial label, which was estimated by the teacher. We observe that images in the same pair of rows contain semantically similar objects.}
\label{fig:grouped_images}
\vskip -0.2in
\end{figure*}

\begin{table}[t]
\caption{{\bf Ablation study on the unlabeled CIFAR-10.} 
We show that both self-labeling and self-attention are necessary to improve the GAN training.}
\label{tbl:ablation_unlabeled}
\begin{center}
\begin{small}
\begin{sc}
\begin{tabular}{@{\hspace{0em}}c@{\hspace{1em}}c@{\hspace{1em}}crr@{\hspace{0em}}}
\toprule
\multirow{2}{*}{}& Self- & Self- & \multirow{2}{*}{\thead[l]{FID-5ep}} & \multirow{2}{*}{\thead[l]{\textoverline{FID}}}\\
& Labeling & Attention & & \\
\midrule
(a)&         - &          - & $7.45 \pm 0.17$ & $6.96 \pm 0.20$\\
(b)&\checkmark &          - & $7.74 \pm 0.20$ & $7.00 \pm 0.34$\\
(c)&\checkmark & \checkmark & $7.28 \pm 0.11$ & $6.81 \pm 0.13$\\
\bottomrule
\end{tabular}
\end{sc}
\end{small}
\end{center}
\vskip -0.1in
\end{table}

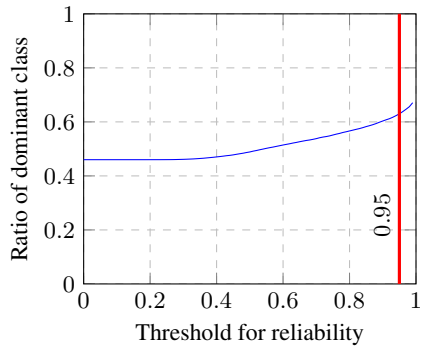
\begin{figure}[t]
\centering
\begin{tikzpicture}
\small
\begin{axis}[
    xlabel={Threshold for reliability},
    ylabel={Ratio of dominant class},
    xmin=0, xmax=1,
    ymin=0, ymax=1,
    width=60mm,
    xtick={0,0.2,0.4,0.6,0.8,1},
    ytick={0,0.2,0.4,0.6,0.8,1},
    ymajorgrids=true,
    xmajorgrids=true,
    grid style=dashed,
]
\addplot[
    color=blue,
    mark=none,
    ]
    coordinates {
(0,0.46004)
(0.01,0.46004)
(0.02,0.46004)
(0.03,0.46004)
(0.04,0.46004)
(0.05,0.46004)
(0.06,0.46004)
(0.07,0.46004)
(0.08,0.46004)
(0.09,0.46004)
(0.1,0.46004)
(0.11,0.46004)
(0.12,0.46004)
(0.13,0.46004)
(0.14,0.46004)
(0.15,0.46004)
(0.16,0.46004)
(0.17,0.46004)
(0.18,0.46004)
(0.19,0.46004)
(0.2,0.460068)
(0.21,0.460084)
(0.22,0.460094)
(0.23,0.460112)
(0.24,0.460176)
(0.25,0.460238)
(0.26,0.460299)
(0.27,0.460451)
(0.28,0.460721)
(0.29,0.461028)
(0.3,0.461344)
(0.31,0.4618)
(0.32,0.462337)
(0.33,0.462891)
(0.34,0.46368)
(0.35,0.464502)
(0.36,0.465383)
(0.37,0.466593)
(0.38,0.467762)
(0.39,0.468931)
(0.4,0.470339)
(0.41,0.471671)
(0.42,0.47318)
(0.43,0.474835)
(0.44,0.476094)
(0.45,0.477879)
(0.46,0.480037)
(0.47,0.482116)
(0.48,0.484099)
(0.49,0.486596)
(0.5,0.488702)
(0.51,0.491669)
(0.52,0.494041)
(0.53,0.49686)
(0.54,0.499405)
(0.55,0.502123)
(0.56,0.504594)
(0.57,0.506799)
(0.58,0.50937)
(0.59,0.511868)
(0.6,0.51444)
(0.61,0.516779)
(0.62,0.519069)
(0.63,0.521745)
(0.64,0.523969)
(0.65,0.526585)
(0.66,0.528966)
(0.67,0.53125)
(0.68,0.533042)
(0.69,0.535377)
(0.7,0.538005)
(0.71,0.541299)
(0.72,0.543811)
(0.73,0.545718)
(0.74,0.548164)
(0.75,0.551418)
(0.76,0.554595)
(0.77,0.557319)
(0.78,0.560806)
(0.79,0.563478)
(0.8,0.566454)
(0.81,0.569421)
(0.82,0.572449)
(0.83,0.575257)
(0.84,0.578719)
(0.85,0.582588)
(0.86,0.585924)
(0.87,0.589626)
(0.88,0.593822)
(0.89,0.598773)
(0.9,0.603277)
(0.91,0.607739)
(0.92,0.611779)
(0.93,0.617071)
(0.94,0.623254)
(0.95,0.629795)
(0.96,0.637984)
(0.97,0.646776)
(0.98,0.656269)
(0.99,0.671047)
    };

\addplot[color=red, style=very thick]
    coordinates{(0.95,0)(0.95,1)}
    node [color=black, above, sloped, pos = .3] {$0.95\quad$};
\end{axis}
\end{tikzpicture}
\vskip -0.2in
\caption{{\bf On the role of self-attention.} The plot shows the average ratio of the dominant class contained in the set of selected real images grouped through the artificial labels. A high rate means that each artificial label is consistent with one real label on average.}
\label{fig:threshold_ave}
\vskip -0.2in
\end{figure}

\begin{figure*}[t]
\begin{center}
\centering
\begin{subfigure}[b]{0.4\linewidth}
\centering
\includegraphics[]{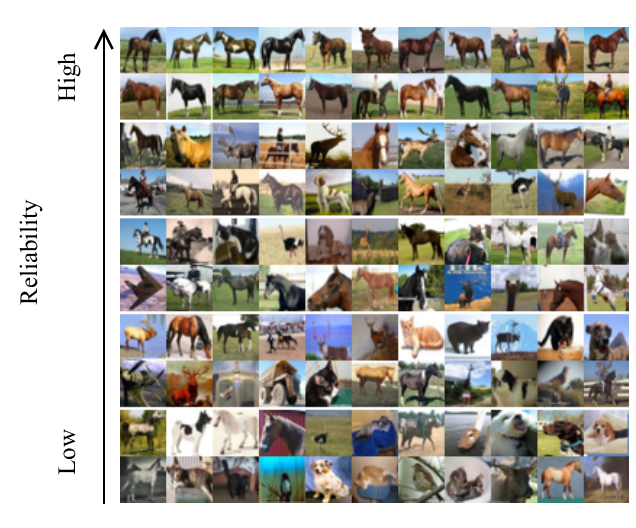}
\vskip -0.1in
\caption{}\label{fig:reliability_p1}
\end{subfigure}
\hspace{1cm}
\begin{subfigure}[b]{0.4\linewidth}
\centering
\begin{tikzpicture}
\small
\begin{axis}[
    xlabel={Reliability threshold (Th)},
    ylabel={Supervised class distribution},
    xmin=0, xmax=1,
    ymin=0, ymax=1,
    width=65mm,
    xtick={0,0.2,0.4,0.6,0.8,1},
    ytick={0,0.2,0.4,0.6,0.8,1},
    ymajorgrids=true,
    xmajorgrids=true,
    grid style=dashed,
    stack plots=y,%
    area style,
]
\addplot[color=black, fill=blue!20] 
table [x=th, y=airplane, col sep=comma] {plt/th_chart_p1.txt}
\closedcycle;
 
\addplot[color=black, fill=red!20] 
table [x=th, y=automobile, col sep=comma] {plt/th_chart_p1.txt}
\closedcycle;
 
\addplot[color=black, fill=yellow!20] 
table [x=th, y=bird, col sep=comma] {plt/th_chart_p1.txt}
node [color=black, below, sloped, pos = .1] {\small Bird}
\closedcycle;

\addplot[color=black, fill=green!20] 
table [x=th, y=cat, col sep=comma] {plt/th_chart_p1.txt}
node [color=black, below, sloped, pos = .1] {\small Cat} 
\closedcycle;;

\addplot[color=black, fill=orange!20] 
table [x=th, y=deer, col sep=comma] {plt/th_chart_p1.txt}
node [color=black, below, sloped, pos = .1] {\small Deer} 
\closedcycle;;

\addplot[color=black, fill=lime!20] 
table [x=th, y=dog, col sep=comma] {plt/th_chart_p1.txt}
node [color=black, below, sloped, pos = .1] {\small Dog} 
\closedcycle;;

\addplot[color=black, fill=gray!20] 
table [x=th, y=frog, col sep=comma] {plt/th_chart_p1.txt}
\closedcycle;

\addplot[color=black, fill=pink!20] 
table [x=th, y=horse, col sep=comma] {plt/th_chart_p1.txt}
node [color=black, below, sloped, pos = .1] {\small Horse} 
\closedcycle;;

\addplot[color=black, fill=purple!20] 
table [x=th, y=ship, col sep=comma] {plt/th_chart_p1.txt}
\closedcycle;

\addplot[color=black, fill=cyan!20] 
table [x=th, y=truck, col sep=comma] {plt/th_chart_p1.txt}
\closedcycle;

\end{axis}
\end{tikzpicture}
\vskip -0.2in
\caption{}\label{fig:threshold_p1}
\end{subfigure}
\begin{subfigure}[b]{0.4\linewidth}
\centering
\includegraphics[]{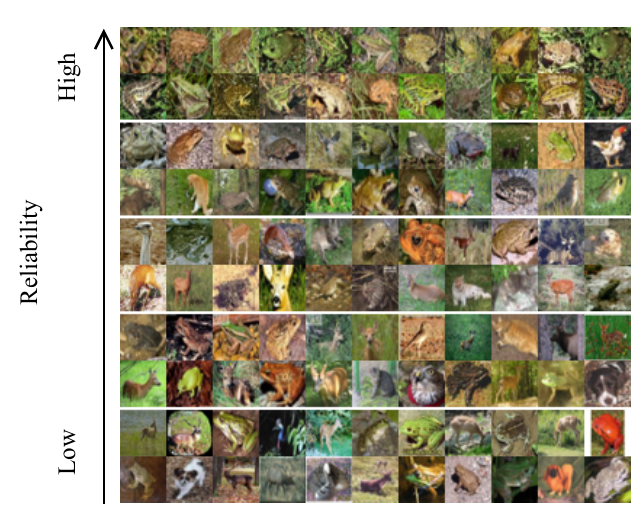}
\vskip -0.1in
\caption{}\label{fig:reliability_p6}
\end{subfigure}
\hspace{1cm}
\begin{subfigure}[b]{0.4\linewidth}
\centering
\begin{tikzpicture}
\small
\begin{axis}[
    xlabel={Reliability threshold (Th)},
    ylabel={Supervised class distribution},
    xmin=0, xmax=1,
    ymin=0, ymax=1,
    width=65mm,
    xtick={0,0.2,0.4,0.6,0.8,1},
    ytick={0,0.2,0.4,0.6,0.8,1},
    ymajorgrids=true,
    xmajorgrids=true,
    grid style=dashed,
    stack plots=y,%
    area style,
]
\addplot[color=black, fill=blue!20] 
table [x=th, y=airplane, col sep=comma] {plt/th_chart_p6.txt}
\closedcycle;
 
\addplot[color=black, fill=red!20] 
table [x=th, y=automobile, col sep=comma] {plt/th_chart_p6.txt}
\closedcycle;
 
\addplot[color=black, fill=yellow!20] 
table [x=th, y=bird, col sep=comma] {plt/th_chart_p6.txt}
node [color=black, below, sloped, pos = .1] {\small Bird}
\closedcycle;

\addplot[color=black, fill=green!20] 
table [x=th, y=cat, col sep=comma] {plt/th_chart_p6.txt}
\closedcycle;;

\addplot[color=black, fill=orange!20] 
table [x=th, y=deer, col sep=comma] {plt/th_chart_p6.txt}
node [color=black, below, sloped, pos = .1] {\small Deer} 
\closedcycle;;

\addplot[color=black, fill=lime!20] 
table [x=th, y=dog, col sep=comma] {plt/th_chart_p6.txt}
\closedcycle;;

\addplot[color=black, fill=gray!20] 
table [x=th, y=frog, col sep=comma] {plt/th_chart_p6.txt}
node [color=black, below, sloped, pos = .1] {\small Frog} 
\closedcycle;

\addplot[color=black, fill=pink!20] 
table [x=th, y=horse, col sep=comma] {plt/th_chart_p6.txt}
\closedcycle;;

\addplot[color=black, fill=purple!20] 
table [x=th, y=ship, col sep=comma] {plt/th_chart_p6.txt}
\closedcycle;

\addplot[color=black, fill=cyan!20] 
table [x=th, y=truck, col sep=comma] {plt/th_chart_p6.txt}
\closedcycle;

\end{axis}
\end{tikzpicture}
\vskip -0.2in
\caption{}\label{fig:threshold_p6}
\end{subfigure}
\vskip -0.1in
\caption{
{\bf Two examples of the reliability of the artificial labels.} (a) \& (c): Pairs of rows with samples of real images assigned to \textit{a single artificial label} and selected via self-attention (for reliability thresholds $Th=0~ (\text{bottom}), 0.25, 0.5, 0.75, \text{and } 1~ (\text{top})$). The top and bottom rows correspond to high and low reliability of the artificial labels respectively. The top row pair contains images that are more consistent with a single real class (\texttt{Horse} in (a) and (b), and \texttt{Frog} in (c) and (d)) than images in the bottom row pair. (b) \& (d): The relative distribution of real classes on images from (a) and (c) respectively. The ratio of the dominant real class (\texttt{Horse} in (b) and \texttt{Frog} in (d)) grows as the threshold for the reliability $p_c$ increases. This shows a desirable alignment between the artificial labels and the real labels especially when the threshold $Th$ is high.}\label{fig:reliability}
\end{center}
\vskip -0.2in
\end{figure*}
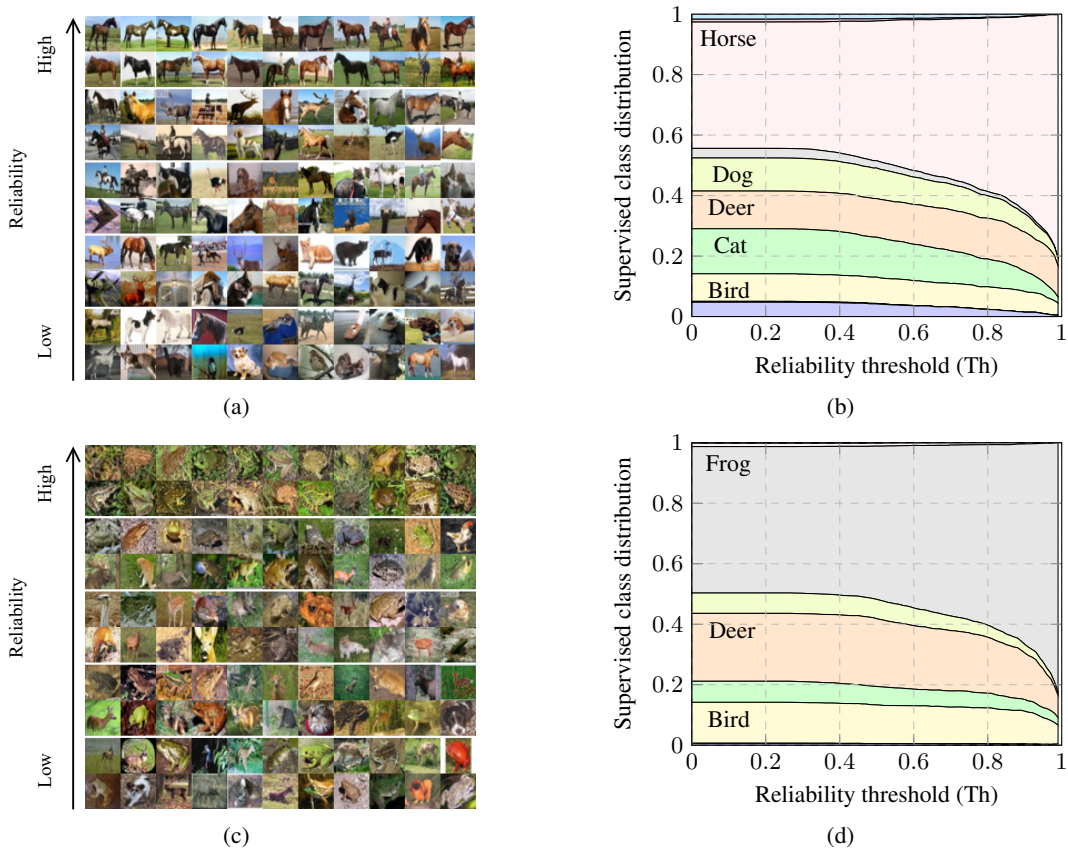

\begin{table}[t]
\caption{Comparison on the unlabeled CIFAR-10.}
\label{tbl:SOTA_GAN}
\begin{center}
\begin{small}
\begin{sc}
\begin{tabular}{lr}
\toprule
Method & \thead[c]{\textoverline{FID}}   \\
\midrule
BigGAN \cite{biggan}          & $14.73$ \\
SS-GAN \cite{self_supervised_gan} & $15.60$ \\
MS-GAN \cite{ssgan_minimax}   & $11.40$ \\
Top-k GAN \cite{ktop_gan}     & $13.34$ \\
ICR-GAN \cite{icr_gan}        & $9.21$ \\
slcGAN \cite{slcgan}          & $8.95$ \\
Top-k ICR-GAN \cite{ktop_gan} & $8.57$ \\
Ours   & $\mathbf{6.81}$ \\
\bottomrule
\end{tabular}
\end{sc}
\end{small}
\end{center}
\vskip -0.1in
\end{table}

\begin{table}[t]
\caption{FID on unlabeled CIFAR-10}
\label{tbl:cifar10_k}
\vskip 0.05in
\begin{center}
\begin{small}
\begin{sc}
\begin{tabular}{lrrrrr}
\toprule
    & baseline & K=2 & K=5 & K=10 & K=50  \\
\midrule
FID & 6.96 & 6.07 & {\bf 5.97} & 6.81 & 7.54 \\
\bottomrule
\end{tabular}
\end{sc}
\end{small}
\end{center}
\vskip -0.2in
\end{table}

\begin{figure*}[t]
\centering
\subfloat[Real images]{
\hspace{-1.8mm}
\includegraphics[width=.49\linewidth]{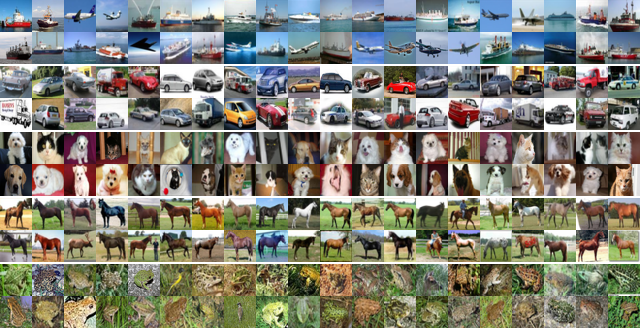}
\label{fig:k5_reals_unlabeled}
}
\subfloat[Fake images]{
\includegraphics[width=.49\linewidth]{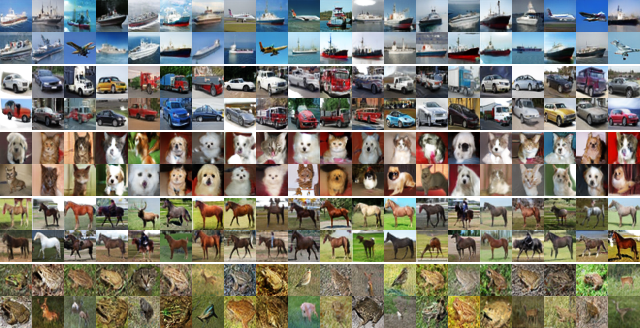}
\label{fig:k5_fakes_unlabeled}
}
\caption{{\bf (a)~Real images and (b)~Fake images on the unlabeled CIFAR-10 with $K=5$ artificial classes, which is half of the actual number of supervised classes.} From the top to the bottom, every pair of image rows corresponds to the same artificial label, which was estimated by the teacher. We observe that the teacher assigns the same artificial label to visually similar objects such as \texttt{Airplane} and \texttt{Ship}, \texttt{Automobile} and \texttt{Truck}, and \texttt{Cat} and \texttt{Dog}.}
\label{fig:grouped_k5_images}
\vskip -0.2in
\end{figure*}

\subsection{Implementation}
{\bf Teacher:} We employ Resnet18~\cite{resnet} with a head of $K$ outputs as the backbone of the teacher.
We use the multi-class cross entropy loss for the classification loss.
The strong image augmentation function $A$ is RandAugment~\cite{randaugment} and the weak image augmentation function $\alpha$ consists of random horizontal-flips and shifts between $\pm4$ pixels. 

{\bf Conditional GAN:} We employ BigGAN~\cite{biggan} for the backbones of the generator (to which we add the conditional label input) and the discriminator.
We also add a fully connected layer of $2K$ outputs and a U-Net decoder~\cite{unet_gan} of $W\times H$ outputs, \emph{i.e.}, the same size of the training image after the global pooling layer of the discriminator, as the heads for conditional adversarial loss and unconditional adversarial loss respectively.
We use the hinge loss for the unconditional adversarial loss and the multi-class hinge loss for the conditional adversarial loss.
To stabilize the training, we apply the following GAN training techniques: differentiable augmentation~\cite{diffaug_gan}, R1 gradient penalty~\cite{GP_r1}, spectral normalization~\cite{sn_gan}, and the U-Net CutMix consistency regularization~\cite{unet_gan}.

We show the pseudo code of the training in \autoref{alg:ours}.
The parameters of the generator $\theta_G$, the discriminator $\theta_D$, and the teacher $\theta_C$ are initialized with Xavier uniform initialization~\cite{xavier_init}.
Firstly, we update the parameters $\theta_C$ of the teacher and assign the artificial labels for the real images.
Secondly, we update the parameters of the cGAN ($\theta_G$ and $\theta_D$) by using the artificial labels obtained in the first step.
$D^U$ and $D^C$ represent the discriminator's heads for the unconditional/conditional adversarial loss.
Then, we repeat the above 2 steps for the number of training iterations.

\section{Experiments}

We evaluate our method on CIFAR-10, STL-10, and SVHN by using FID scores as a quantitative measure and also visualize samples for a qualitative assessment.
The FID scores are computed by using the official implementation~\cite{tf_fid} on $50{,}000$ generated samples.
Moreover, we use the FID-5ep, which is the FID averaged over the last 5 epochs of 5 evaluation runs and also the \textoverline{FID}, which is the average of the lowest FIDs in 5 evaluation runs.
We train the network on a single GPU: GeForce RTX 2080ti for CIFAR-10 and SVHN and Quadro RTX 6000 for STL-10.

To train the teacher we use Nesterov's momentum optimizer with a batch size of $64$, momentum $0.9$, $K=10$, EMA decay of $0.999$, and the number of epochs is $40$ and $60$ on the unlabeled and labeled datasets respectively.
To train the cGAN we use Adam's optimizer with a batch size of $128$ for the loss with the artificial labels and of $64$ for the other losses, $Th=0.95$, and the gradient penalty weight is $10$.
We use the large batch size for the unlabeled real images, because we select a subset via self-attention.

\begin{table}[t]
\caption{{\bf Ablation study on the labeled CIFAR-10.} We calculate the conditional adversarial loss with (\textsc{a}) the artificial labels, (\textsc{b}) the real labels, and (\textsc{c}) both labels. 
}
\label{tbl:ablation_labeled}
\begin{center}
\begin{small}
\begin{sc}
\begin{tabular}{ccrr}
\toprule
& Labels &  \thead[l]{FID-5ep} & \thead[l]{\textoverline{FID}}\\
\midrule
(a) & Artificial       &  $7.28 \pm 0.11$ & $6.81 \pm 0.13$\\
(b) & Real            &  $5.04 \pm 0.08$ & $4.57 \pm 0.10$\\
(c) & Artificial \& Real &  $4.72 \pm 0.06$ & $4.35 \pm 0.05$\\
\bottomrule
\end{tabular}
\end{sc}
\end{small}
\end{center}
\vskip -0.1in
\end{table}

\begin{figure}[t]
\begin{center}
\begin{tikzpicture}
\small
\begin{axis}[
    xlabel={Rate of labeled samples [\%]},
    ylabel={FID},
    xmin=0, xmax=100,
    ymin=0, ymax=9,
    width=60mm,
    xtick={0,25,50,75,100},
    ytick={0,1,2,3,4,5,6,7,8,9},
    ymajorgrids=true,
    xmajorgrids=true,
    grid style=dashed,
]
\addplot[
    color=blue,
    mark=square,
    ]
    coordinates {
    (0,6.81)(25,5.83)(50,6.276)(75,5.91)(100,4.35)
    };
\end{axis}
\end{tikzpicture}
\vskip -0.2in
\caption{{\bf FID for different real label rates (in percent with respect to the complete set of labels) in CIFAR-10.} Our method can handle different levels of labeling in a unified manner. We find that the FID tends to improve as more real labels are available.}
\label{fig:ablation_semi_labeled}
\end{center}
\vskip -0.2in
\end{figure}
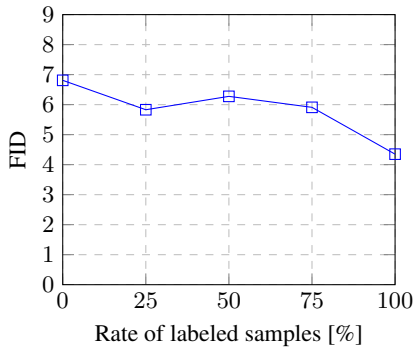

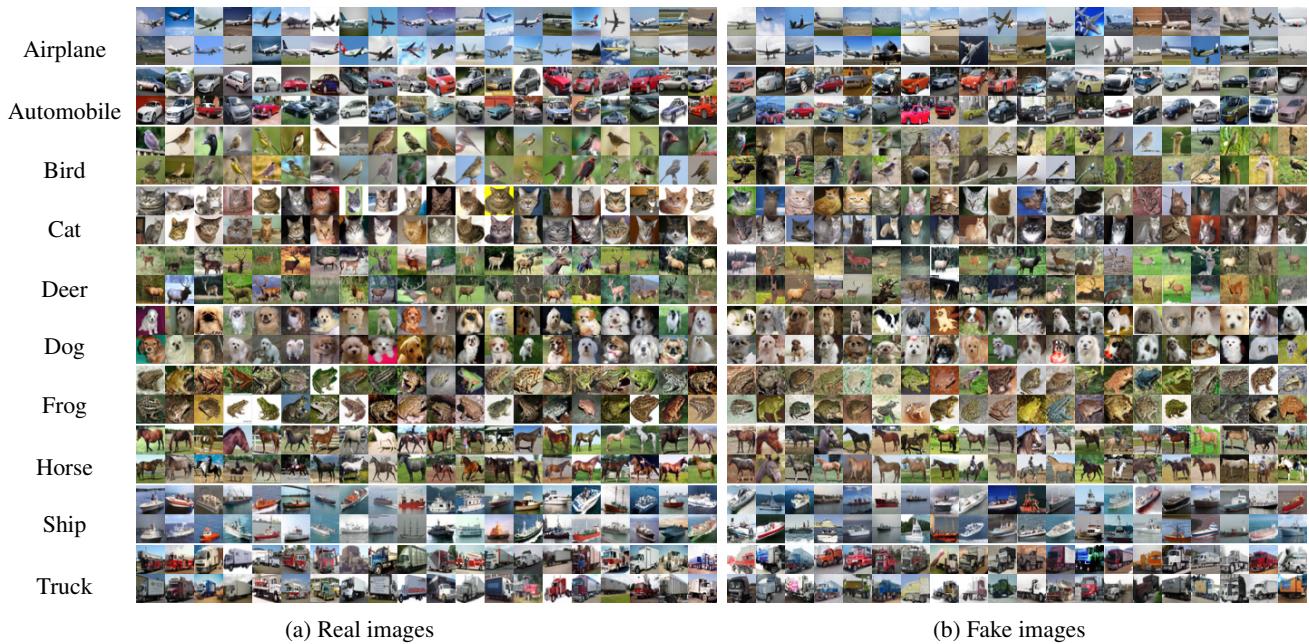
\begin{figure*}[t]
\begin{center}
\centering
\hspace{-5mm}
\begin{subfigure}[b]{96mm}
\begin{tikzpicture}
\def\ccheight{79mm}
\node[] at (0,0.9*\ccheight) {\small Airplane};
\node[] at (0,0.8*\ccheight) {\small Automobile};
\node[] at (0,0.7*\ccheight) {\small Bird};
\node[] at (0,0.6*\ccheight) {\small Cat};
\node[] at (0,0.5*\ccheight) {\small Deer};
\node[] at (0,0.4*\ccheight) {\small Dog};
\node[] at (0,0.3*\ccheight) {\small Frog};
\node[] at (0,0.2*\ccheight) {\small Horse};
\node[] at (0,0.1*\ccheight) {\small Ship};
\node[] at (0,0.0*\ccheight) {\small Truck};
\end{tikzpicture}
\includegraphics[width=77mm]{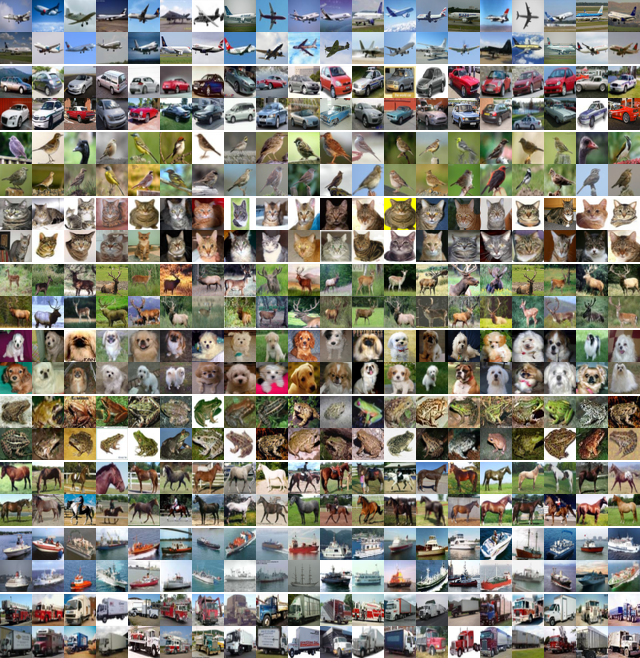}
\caption{Real images}\label{fig:reals_labeled}
\end{subfigure}
\begin{subfigure}[b]{75mm}
\centering
\includegraphics[width=77mm]{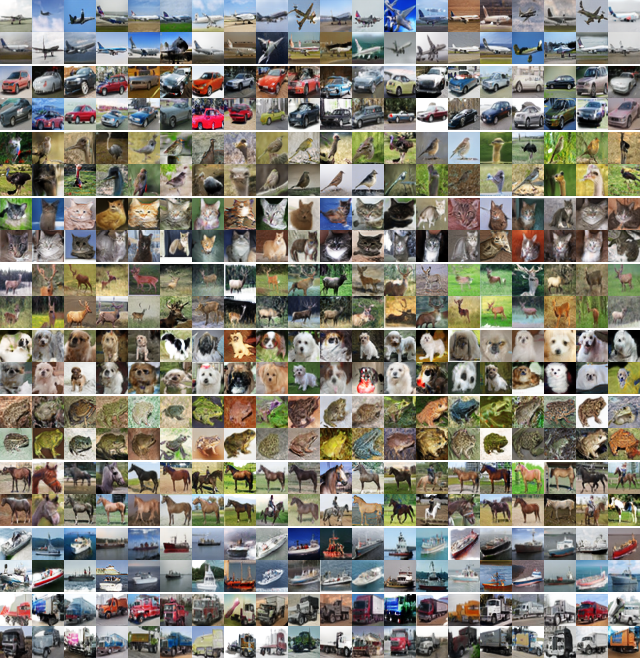}
\caption{Fake images}\label{fig:fakes_labeled}
\end{subfigure}
\vskip -0.1in
\caption{{\bf (a)~Real images and (b)~Fake images on the labeled CIFAR-10.} 
From the top to the bottom, every pair of image rows corresponds to the same artificial label, which was estimated by the teacher. Images in the same pair of rows contain semantically similar objects and are aligned with the real labels (the real labels corresponding to the same artificial label index are shown on the left).}
\label{fig:grouped_images_labeled}
\end{center}
\vskip -0.2in
\end{figure*}

\begin{table}[t]
\caption{Comparison on the labeled CIFAR-10.}
\label{tbl:SOTA_cGAN}
\begin{center}
\begin{small}
\begin{sc}
\begin{tabular}{lr}
\toprule
Method & \thead[c]{\textoverline{FID}}  \\
\midrule
BigGAN \cite{biggan}            & $9.06$ \\
DiffAug GAN \cite{diffaug_gan}  & $8.56$ \\
MHinge GAN \cite{mhinge_gan}    & $6.40$ \\
FQ-GAN \cite{fq_gan}            & $5.39$ \\
Ours                            & $\mathbf{4.35}$ \\
\bottomrule
\end{tabular}
\end{sc}
\end{small}
\end{center}
\vskip -0.1in
\end{table}

\begin{table}[t]
\caption{{\bf FID on STL-10.} The STL-10 dataset contains $100{,}000$ unlabeled images and $500$ labeled images.}
\label{tbl:ablation_stl}
\begin{center}
\begin{small}
\begin{sc}
\begin{tabular}{lrr}
\toprule
Dataset &\thead[l]{Baseline} & \thead[l]{+Ours}\\
\midrule
Unlabeled & $32.85$ & $30.91$\\
$+0.5\%$ labeled & $48.49$ & $43.29$ \\
\bottomrule
\end{tabular}
\end{sc}
\end{small}
\end{center}
\vskip -0.2in
\end{table}

\begin{table}[t]
\caption{{\bf Comparison on the unlabeled STL-10.} The methods below do not use the same generator, but the number of the convolutional layers is the same. Our FID is comparable to that of other methods, although we did not use network architecture search.}
\label{tbl:SOTA_STL}
\begin{center}
\begin{small}
\begin{sc}
\begin{tabular}{lS}
\toprule
Method & {\thead[c]{FID}}   \\
\midrule
SNGAN \cite{sn_gan}      & 40.1 \\
AutoGAN \cite{auto_gan}  & 31.01 \\
DEGAS \cite{degas}       & 28.76 \\
Ours      & 30.91 \\
\bottomrule
\end{tabular}
\end{sc}
\end{small}
\end{center}
\vskip -0.2in
\end{table}

\begin{table}[t]
\caption{{\bf FID on the unlabeled STL-10 and SVHN.} Our method improves the FID scores on both datasets.}
\label{tbl:stl_svhn_unlabeled}
\begin{center}
\begin{small}
\begin{sc}
\begin{tabular}{lrr}
\toprule
Dataset &\thead[l]{Baseline} & \thead[l]{+Ours}\\
\midrule
STL-10 & $32.85$ & $30.91$\\
SVHN & $2.44$ & $2.19$ \\
\bottomrule
\end{tabular}
\end{sc}
\end{small}
\end{center}
\vskip -0.2in
\end{table}

\begin{figure*}[t]
\begin{center}
\subfloat[Real images]{
\hspace{-1.8mm}
\includegraphics[width=.49\linewidth]{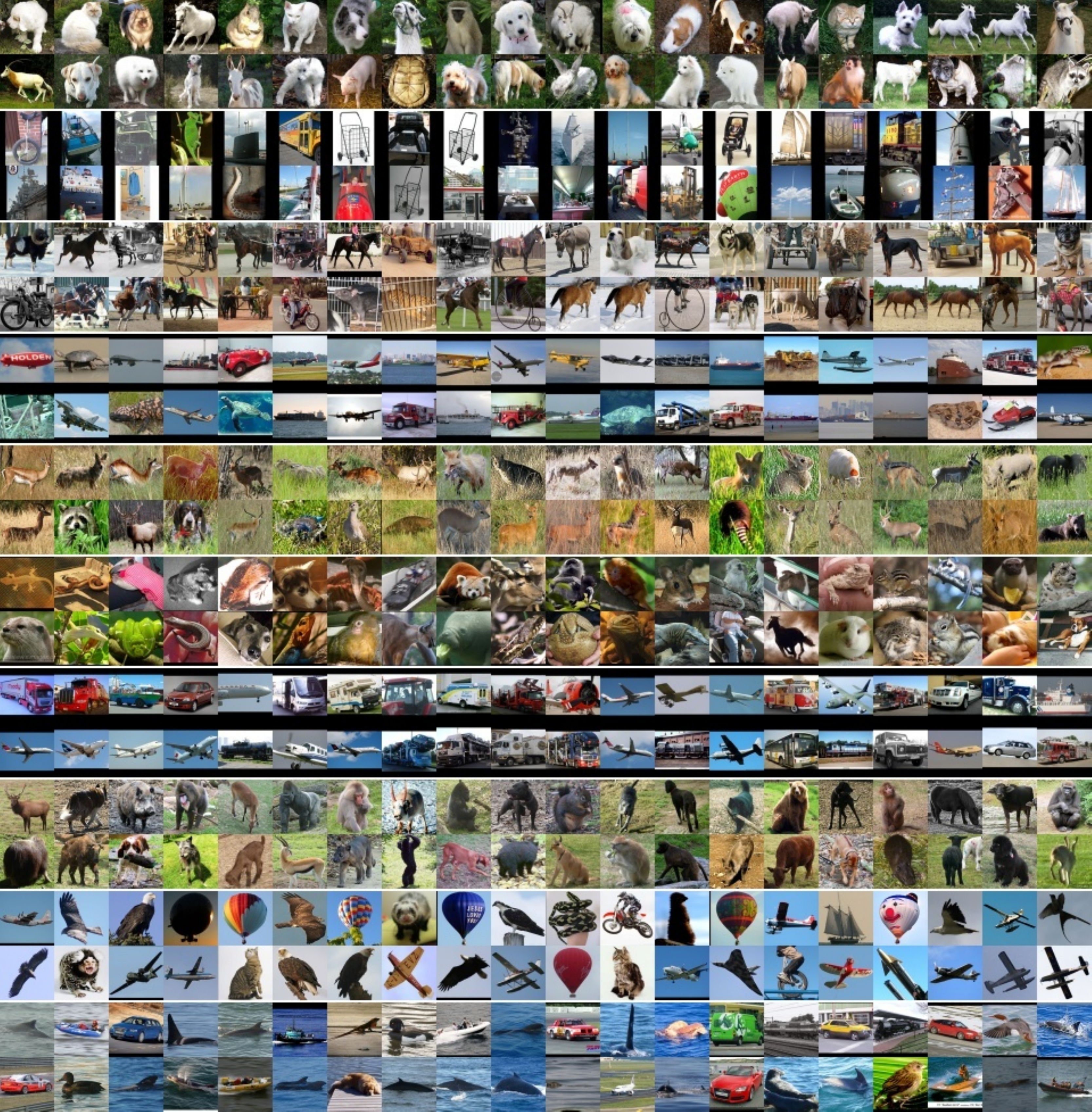}
\label{fig:stl_reals_unlabeled}
}
\subfloat[Fake images]{
\includegraphics[width=.49\linewidth]{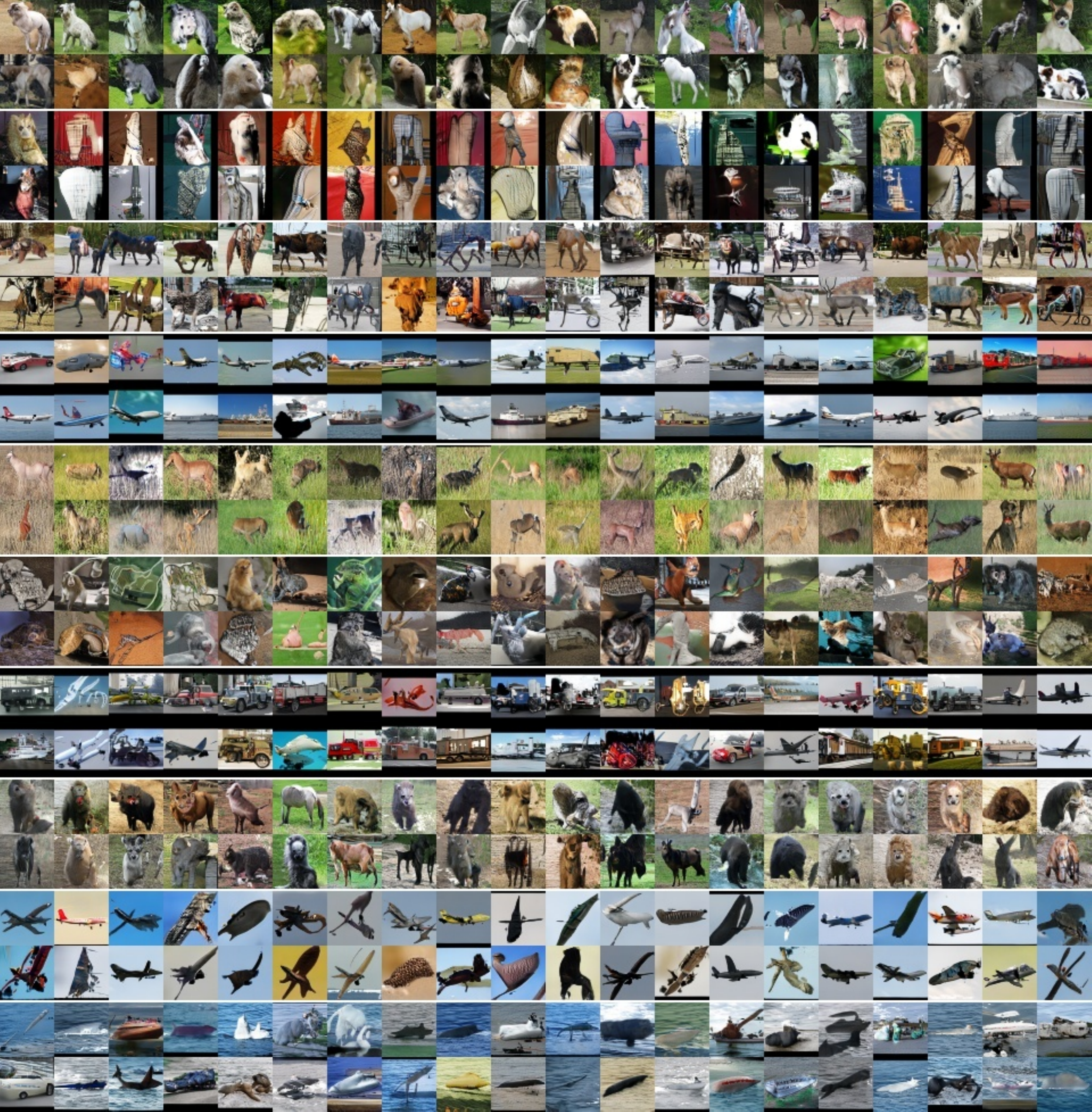}
\label{fig:stl_fakes_unlabeled}
}
\caption{{\bf (a)~Real images and (b)~Fake images on the unlabeled STL-10.} 
From the top to the bottom, every pair of image rows corresponds to the same artificial label, which was estimated by the teacher. As with CIFAR-10, we observe that images in the same pair of rows contain semantically similar objects.}
\label{fig:stl_grouped_images}
\end{center}
\vskip -0.2in
\end{figure*}

\noindent\textbf{Unlabeled CIFAR-10. }
We evaluate the effectiveness of self-labeling and self-attention on CIFAR-10 and summarize the results in \autoref{tbl:ablation_unlabeled}.
The results show that (\textsc{b}) unreliable artificial labels hurt the performance and (\textsc{c}) refined artificial labels help the training of the GAN compared to using (\textsc{a}) no artificial labels.
Our method improves the FID score averaged over last 5 epochs from $7.45$ to $7.28$ and the best FID score averaged over 5 runs from $6.96$ to $6.81$.

In \autoref{fig:reals_unlabeled}, we show real images grouped by their artificial labels, which were learned without supervision. The images are selected via self-attention. By starting from the top, every pair of rows corresponds to one artificial label. We can see that every artificial label identifies images with similar objects, but also that objects across separate labels differ substantially.
For example, the first, the third and the last groups contain an object on the ground, in the sea, and in the sky respectively, and the third and seventh groups contain ships and cars respectively.
In \autoref{fig:fakes_unlabeled}, we also show in the same manner images generated using the artificial labels (as input to the generator). Notice the broad diversity and the strong similarity between fake and real images in terms of artificial categories.

To explain the role of the proposed reliability measure $p_c$ (see \cref{eq:pc}) for self-attention, we sort real images with the same dominant artificial label based on the magnitude of the reliability. We show in \autoref{fig:reliability} an evaluation of the consistency between real and artificial labels for two randomly chosen artificial labels. Through visual inspection we find that these labels correspond mostly to the \texttt{Horse} (\autoref{fig:reliability_p1}) and \texttt{Frog} (\autoref{fig:reliability_p6}) categories. 
The top and bottom rows correspond to images of high and low reliability respectively.
One can observe the higher semantic class consistency (\emph{i.e.}, more \texttt{Horse} images in \autoref{fig:reliability_p1} and more \texttt{Frog} images in \autoref{fig:reliability_p6}), when the reliability is high.
For a more quantitative measure of this consistency, in \autoref{fig:threshold_p1} and \autoref{fig:threshold_p6} we use labeled data to show the relative distribution of real samples selected with different reliability thresholds ($Th$).
We notice that the ratio of the dominant classes \texttt{Horse} and \texttt{Frog} grows as the threshold is increased, which suggests a consistency between the clustering induced by artificial labels and real labels as well as a correlation between the reliability $p_c$ and the purity measure of the selected samples.
Moreover, in \autoref{fig:threshold_ave}, we show the rate of the dominant class averaged over all artificial labels.
The plot shows that the ratio of the dominant class (that is consistent with a manually defined class) grows with the reliability threshold. We also indicate with a red vertical bar the chosen threshold in our implementation. The threshold defines a trade-off between high class consistency and number of selected real samples. As also shown previously in \autoref{tbl:ablation_unlabeled}, this selection process is quite essential to self-labeling.

Finally, we compare our method with the state-of-the-art methods for unsupervised GAN training on CIFAR-10 in \autoref{tbl:SOTA_GAN}.
The methods use different loss functions, but share the same BigGAN generator.
Although our generator uses the conditional label input, the basic backbone is the same.
Our proposed training shows a significant \textoverline{FID} improvement over the previous state-of-the-art (from $8.57$ to $6.81$).

To evaluate the effect of the number of the artificial classes $K$, in \autoref{tbl:cifar10_k} we show the results learned with several settings of $K$ on CIFAR-10.
The best FID $5.97$ is obtained with $K=5$, which is smaller than the number of real classes~($10$).
In \autoref{fig:grouped_k5_images}, we show real and fake images learned with $K=5$, similarly to what shown in \autoref{fig:grouped_images}.
We can see that real classes that are difficult to tell apart are grouped together. For example, the teacher clusters the following pairs \texttt{Airplane} and \texttt{Ship}, \texttt{Automobile} and \texttt{Truck}, and \texttt{Cat} and \texttt{Dog}.
We find that the conﬁdence of the teacher on artificial labels is higher with $K=5$ than with $K=10$, which is the true number of real classes. 
Also, there is much less conﬁdence with $K=50$.
This further supports the argument that reducing the ambiguities of the artificial label helps the discriminator, and then in turn the generator.
The artificial labels contain less information with $K=2$ than with $K=5$, but the difference of the FID is small.
The result suggests that a large $K$ is suitable as long as the estimation of the artificial labels is confident.

\noindent\textbf{Labeled CIFAR-10. }
As shown in~\autoref{tbl:ablation_labeled}, our method also improves the FID score when training on labeled datasets.
The first column shows the type of labels used in the conditional adversarial loss.
We calculate the conditional adversarial loss with either (a)~the artificial labels, (b)~the real labels, or (c)~both of them. The result using both labels yields the best FID.
Our method improves the baseline cGAN on the FID averaged on the last 5 epochs from $5.04$ to $4.72$ and the best \textoverline{FID} from $4.57$ to $4.35$.
The results show that the artificial labels integrate naturally with the real labels and further boost the performance of the generator.

In \autoref{fig:ablation_semi_labeled}, we also evaluate our method in the semi-supervised learning settings, by using a partially labeled CIFAR-10 dataset.
We randomly select $25\%$, $50\%$, and $75\%$ of the available real labels to train the cGAN with our method.
The rates of $0\%$ and $100\%$ mean unlabeled and fully labeled respectively.
We observe that the FID tends to improve with an increasing number of real labels. 

In~\autoref{fig:grouped_images_labeled}, we show the real and generated images grouped by the artificial labels $\hat{y}_c$ (estimated by the teacher) and the artificial conditional labels $c$ respectively.
\autoref{fig:reals_labeled} shows qualitatively that every group corresponds to some meaningful real class although the teacher is trained only on fake images. 
\autoref{fig:fakes_labeled} shows that the conditional labels $c$ seem to seamlessly align with the available real labels.
To measure the degree of alignment between artificial and real labels, we compute the classification accuracy on test real images after finding the optimal correspondence between the predicted and ground truth labels. We find that the accuracy of the classifier averaged over 5 runs is $43\%$ in the unsupervised training case and $83\%$ in the fully supervised training case.
We attribute this alignment to the fact that artificial labels are arbitrarily defined by the generator and the generator can adapt to the backpropagation from the discriminator, which in turn is influenced by the real labels.

In \autoref{tbl:SOTA_cGAN}, we compare our method with the state-of-the-art cGAN methods on CIFAR-10. As in the unsupervised case, our proposed training shows a significant \textoverline{FID} improvement over the previous state-of-the-art (from $5.39$ to $4.35$).

\noindent\textbf{Evaluation on STL-10 and SVHN. }
In \autoref{tbl:stl_svhn_unlabeled}, we compare our method to a baseline without self-labeling and self-attention on STL-10 and SVHN. As we can see from the range of the FID scores, the STL-10 dataset is more complex and the SVHN dataset is simpler than the CIFAR-10 dataset. Another difference is that we use an image size of $48\times48$ pixels for the experiments on STL-10, while in the CIFAR-10 and SVHN datasets it is of $32\times32$ pixels.
The results show that our method improves the FID score on both datasets compared to the baseline (from $32.85$ to $30.91$ and from $2.44$ to $2.19$). In \autoref{fig:stl_grouped_images}, we show samples of real and generated images.

In \autoref{tbl:ablation_stl}, we evaluate our method on STL-10 in the semi-supervised setting.
The STL-10 dataset contains $100{,}000$ images of various classes and $500$ images of $10$ classes with supervised labels.
By adding extremely few real labels to the cGAN training, the FID score of the baseline worsens (from $32.85$ to $48.49$). However, our method improves the FID score with respect to the baseline and even more substantially when labels are available (from $48.49$ to $43.29$).  
Finally, in \autoref{tbl:SOTA_STL}, we compare our method with the state-of-the-art unsupervised GANs.
We should point out that these methods use different generators, as a result of network architecture search~\cite{auto_gan, degas}. Despite the non optimality of our generator architecture, our approach yields a very competitive FID.

\section{Conclusions}

We proposed a novel GAN training scheme that can handle different levels of labeling in a unified manner.
Our approach is based on using the generator to implicitly define artificial labels and then to train a classifier on purely synthetic data and labels. This classifier can then be used to self-label real data. Its class-consistency is found to correlate well with its classification probability score, which we then use to select samples with a reliable label (self-attention).
We evaluated our approach on CIFAR-10, STL-10 and SVHN, and showed that both self-labeling and self-attention consistently improve the quality of generated data.

\bibliography{main}

\begin{thebibliography}{34}
\providecommand{\natexlab}[1]{#1}
\providecommand{\url}[1]{\texttt{#1}}
\expandafter\ifx\csname urlstyle\endcsname\relax
  \providecommand{\doi}[1]{doi: #1}\else
  \providecommand{\doi}{doi: \begingroup \urlstyle{rm}\Url}\fi

\bibitem[Berthelot et~al.(2019)Berthelot, Carlini, Cubuk, Kurakin, Sohn, Zhang,
  and Raffel]{remixmatch}
Berthelot, D., Carlini, N., Cubuk, E.~D., Kurakin, A., Sohn, K., Zhang, H., and
  Raffel, C.
\newblock Remixmatch: Semi-supervised learning with distribution alignment and
  augmentation anchoring.
\newblock \emph{CoRR}, abs/1911.09785, 2019.

\bibitem[Brock et~al.(2019)Brock, Donahue, and Simonyan]{biggan}
Brock, A., Donahue, J., and Simonyan, K.
\newblock Large scale {GAN} training for high fidelity natural image synthesis.
\newblock In \emph{7th International Conference on Learning Representations,
  {ICLR} 2019, New Orleans, LA, USA, May 6-9, 2019}. OpenReview.net, 2019.

\bibitem[Chen et~al.(2019{\natexlab{a}})Chen, Zhai, Ritter, Lucic, and
  Houlsby]{self_supervised_gan}
Chen, T., Zhai, X., Ritter, M., Lucic, M., and Houlsby, N.
\newblock Self-supervised {GANs} via auxiliary rotation loss.
\newblock In \emph{2019 {IEEE} Conference on Computer Vision and Pattern
  Recognition, {CVPR} 2019, Long Beach, CA, USA, June 16-20, 2019}, pp.\
  12154--12163. {IEEE}, 2019{\natexlab{a}}.

\bibitem[Chen et~al.(2019{\natexlab{b}})Chen, Zhai, Ritter, Lucic, and
  Houlsby]{ssgan_rotation}
Chen, T., Zhai, X., Ritter, M., Lucic, M., and Houlsby, N.
\newblock Self-supervised {GANs} via auxiliary rotation loss.
\newblock In \emph{2019 {IEEE} Conference on Computer Vision and Pattern
  Recognition, {CVPR} 2019, Long Beach, CA, USA, June 16-20, 2019}, pp.\
  12154--12163. {IEEE}, 2019{\natexlab{b}}.

\bibitem[Coates et~al.(2011)Coates, Ng, and Lee]{stl10}
Coates, A., Ng, A.~Y., and Lee, H.
\newblock An analysis of single-layer networks in unsupervised feature
  learning.
\newblock In Gordon, G.~J., Dunson, D.~B., and Dud{\'{\i}}k, M. (eds.),
  \emph{Proceedings of the Fourteenth International Conference on Artificial
  Intelligence and Statistics, {AISTATS} 2011, Fort Lauderdale, USA, April
  11-13, 2011}, volume~15 of \emph{{JMLR} Proceedings}, pp.\  215--223.
  JMLR.org, 2011.

\bibitem[Cubuk et~al.(2020)Cubuk, Zoph, Shlens, and Le]{randaugment}
Cubuk, E.~D., Zoph, B., Shlens, J., and Le, Q.
\newblock Randaugment: Practical automated data augmentation with a reduced
  search space.
\newblock In Larochelle, H., Ranzato, M., Hadsell, R., Balcan, M., and Lin, H.
  (eds.), \emph{Advances in Neural Information Processing Systems 33: Annual
  Conference on Neural Information Processing Systems 2020, NeurIPS 2020,
  December 6-12, 2020, virtual}, 2020.

\bibitem[Doveh \& Giryes(2019)Doveh and Giryes]{degas}
Doveh, S. and Giryes, R.
\newblock {DEGAS:} differentiable efficient generator search.
\newblock \emph{CoRR}, abs/1912.00606, 2019.

\bibitem[Glorot \& Bengio(2010)Glorot and Bengio]{xavier_init}
Glorot, X. and Bengio, Y.
\newblock Understanding the difficulty of training deep feedforward neural
  networks.
\newblock In Teh, Y.~W. and Titterington, D.~M. (eds.), \emph{Proceedings of
  the Thirteenth International Conference on Artificial Intelligence and
  Statistics, {AISTATS} 2010, Chia Laguna Resort, Sardinia, Italy, May 13-15,
  2010}, volume~9 of \emph{{JMLR} Proceedings}, pp.\  249--256. JMLR.org, 2010.

\bibitem[Gong et~al.(2019)Gong, Chang, Jiang, and Wang]{auto_gan}
Gong, X., Chang, S., Jiang, Y., and Wang, Z.
\newblock {AutoGAN}: Neural architecture search for generative adversarial
  networks.
\newblock In \emph{2019 {IEEE} International Conference on Computer Vision,
  {ICCV} 2019, Seoul, Korea (South), October 27 - November 2, 2019}, pp.\
  3223--3233. {IEEE}, 2019.

\bibitem[Goodfellow et~al.(2014)Goodfellow, Pouget{-}Abadie, Mirza, Xu,
  Warde{-}Farley, Ozair, Courville, and Bengio]{gan}
Goodfellow, I.~J., Pouget{-}Abadie, J., Mirza, M., Xu, B., Warde{-}Farley, D.,
  Ozair, S., Courville, A.~C., and Bengio, Y.
\newblock Generative adversarial nets.
\newblock In Ghahramani, Z., Welling, M., Cortes, C., Lawrence, N.~D., and
  Weinberger, K.~Q. (eds.), \emph{Advances in Neural Information Processing
  Systems 27: Annual Conference on Neural Information Processing Systems 2014,
  December 8-13 2014, Montreal, Quebec, Canada}, pp.\  2672--2680, 2014.

\bibitem[He et~al.(2016)He, Zhang, Ren, and Sun]{resnet}
He, K., Zhang, X., Ren, S., and Sun, J.
\newblock Deep residual learning for image recognition.
\newblock In \emph{2016 {IEEE} Conference on Computer Vision and Pattern
  Recognition, {CVPR} 2016, Las Vegas, NV, USA, June 27-30, 2016}, pp.\
  770--778. {IEEE}, 2016.

\bibitem[He et~al.(2020)He, Fan, Wu, Xie, and Girshick]{moco}
He, K., Fan, H., Wu, Y., Xie, S., and Girshick, R.~B.
\newblock Momentum contrast for unsupervised visual representation learning.
\newblock In \emph{2020 {IEEE} Conference on Computer Vision and Pattern
  Recognition, {CVPR} 2020, Seattle, WA, USA, June 13-19, 2020}, pp.\
  9726--9735. {IEEE}, 2020.

\bibitem[Heusel et~al.(2017)Heusel, Ramsauer, Unterthiner, Nessler, and
  Hochreiter]{tf_fid}
Heusel, M., Ramsauer, H., Unterthiner, T., Nessler, B., and Hochreiter, S.
\newblock {GANs} trained by a two time-scale update rule converge to a local
  nash equilibrium.
\newblock In Guyon, I., von Luxburg, U., Bengio, S., Wallach, H.~M., Fergus,
  R., Vishwanathan, S. V.~N., and Garnett, R. (eds.), \emph{Advances in Neural
  Information Processing Systems 30: Annual Conference on Neural Information
  Processing Systems 2017, December 4-9, 2017, Long Beach, CA, {USA}}, pp.\
  6626--6637, 2017.

\bibitem[Karras et~al.(2019)Karras, Laine, and Aila]{style_gan}
Karras, T., Laine, S., and Aila, T.
\newblock A style-based generator architecture for generative adversarial
  networks.
\newblock In \emph{2019 {IEEE} Conference on Computer Vision and Pattern
  Recognition, {CVPR} 2019, Long Beach, CA, USA, June 16-20, 2019}, pp.\
  4401--4410. {IEEE}, 2019.

\bibitem[Kavalerov et~al.(2019)Kavalerov, Czaja, and Chellappa]{mhinge_gan}
Kavalerov, I., Czaja, W., and Chellappa, R.
\newblock {cGANs} with multi-hinge loss.
\newblock \emph{CoRR}, abs/1912.04216, 2019.

\bibitem[Krizhevsky \& Hinton(2009)Krizhevsky and Hinton]{cifar10}
Krizhevsky, A. and Hinton, G.
\newblock Learning multiple layers of features from tiny images.
\newblock In \emph{Technical report}. Citeseer, 2009.

\bibitem[Lucic et~al.(2019)Lucic, Tschannen, Ritter, Zhai, Bachem, and
  Gelly]{s3gan}
Lucic, M., Tschannen, M., Ritter, M., Zhai, X., Bachem, O., and Gelly, S.
\newblock High-fidelity image generation with fewer labels.
\newblock In Chaudhuri, K. and Salakhutdinov, R. (eds.), \emph{Proceedings of
  the 36th International Conference on Machine Learning, {ICML} 2019, 9-15 June
  2019, Long Beach, California, {USA}}, volume~97 of \emph{Proceedings of
  Machine Learning Research}, pp.\  4183--4192. {PMLR}, 2019.

\bibitem[Mescheder et~al.(2018)Mescheder, Geiger, and Nowozin]{GP_r1}
Mescheder, L.~M., Geiger, A., and Nowozin, S.
\newblock Which training methods for {GANs} do actually converge?
\newblock In Dy, J.~G. and Krause, A. (eds.), \emph{Proceedings of the 35th
  International Conference on Machine Learning, {ICML} 2018,
  Stockholmsm{\"{a}}ssan, Stockholm, Sweden, July 10-15, 2018}, volume~80 of
  \emph{Proceedings of Machine Learning Research}, pp.\  3478--3487. {PMLR},
  2018.

\bibitem[Mirza \& Osindero(2014)Mirza and Osindero]{mizraOsindero2018}
Mirza, M. and Osindero, S.
\newblock Conditional generative adversarial nets.
\newblock \emph{CoRR}, abs/1411.1784, 2014.

\bibitem[Miyato \& Koyama(2018)Miyato and Koyama]{proj_gan}
Miyato, T. and Koyama, M.
\newblock {cGANs} with projection discriminator.
\newblock In \emph{6th International Conference on Learning Representations,
  {ICLR} 2018, Vancouver, BC, Canada, April 30 - May 3, 2018, Conference Track
  Proceedings}. OpenReview.net, 2018.

\bibitem[Miyato et~al.(2018)Miyato, Kataoka, Koyama, and Yoshida]{sn_gan}
Miyato, T., Kataoka, T., Koyama, M., and Yoshida, Y.
\newblock Spectral normalization for generative adversarial networks.
\newblock In \emph{6th International Conference on Learning Representations,
  {ICLR} 2018, Vancouver, BC, Canada, April 30 - May 3, 2018, Conference Track
  Proceedings}. OpenReview.net, 2018.

\bibitem[Netzer et~al.(2011)Netzer, Wang, Coates, Bissacco, Wu, and Ng]{svhn}
Netzer, Y., Wang, T., Coates, A., Bissacco, A., Wu, B., and Ng, A.~Y.
\newblock Reading digits in natural images with unsupervised feature learning.
\newblock In \emph{NIPS Workshop on Deep Learning and Unsupervised Feature
  Learning}, 2011.

\bibitem[Noroozi(2020)]{slcgan}
Noroozi, M.
\newblock Self-labeled conditional {GANs}.
\newblock \emph{CoRR}, abs/2012.02162, 2020.

\bibitem[Odena et~al.(2017)Odena, Olah, and Shlens]{ac_gan}
Odena, A., Olah, C., and Shlens, J.
\newblock Conditional image synthesis with auxiliary classifier {GANs}.
\newblock In Precup, D. and Teh, Y.~W. (eds.), \emph{Proceedings of the 34th
  International Conference on Machine Learning, {ICML} 2017, Sydney, NSW,
  Australia, 6-11 August 2017}, volume~70 of \emph{Proceedings of Machine
  Learning Research}, pp.\  2642--2651. {PMLR}, 2017.

\bibitem[Reed et~al.(2016)Reed, Akata, Yan, Logeswaran, Schiele, and
  Lee]{reedLee2016}
Reed, S.~E., Akata, Z., Yan, X., Logeswaran, L., Schiele, B., and Lee, H.
\newblock Generative adversarial text to image synthesis.
\newblock In Balcan, M. and Weinberger, K.~Q. (eds.), \emph{Proceedings of the
  33rd International Conference on Machine Learning, {ICML} 2016, New York
  City, NY, USA, June 19-24, 2016}, volume~48 of \emph{{JMLR} Workshop and
  Conference Proceedings}, pp.\  1060--1069. JMLR.org, 2016.

\bibitem[Sch{\"{o}}nfeld et~al.(2020)Sch{\"{o}}nfeld, Schiele, and
  Khoreva]{unet_gan}
Sch{\"{o}}nfeld, E., Schiele, B., and Khoreva, A.
\newblock A u-net based discriminator for generative adversarial networks.
\newblock In \emph{2020 {IEEE} Conference on Computer Vision and Pattern
  Recognition, {CVPR} 2020, Seattle, WA, USA, June 13-19, 2020}, pp.\
  8204--8213. {IEEE}, 2020.

\bibitem[Sinha et~al.(2020)Sinha, Zhao, Goyal, Raffel, and Odena]{ktop_gan}
Sinha, S., Zhao, Z., Goyal, A., Raffel, C., and Odena, A.
\newblock Top-k training of {GANs}: Improving {GAN} performance by throwing
  away bad samples.
\newblock In Larochelle, H., Ranzato, M., Hadsell, R., Balcan, M., and Lin, H.
  (eds.), \emph{Advances in Neural Information Processing Systems 33: Annual
  Conference on Neural Information Processing Systems 2020, NeurIPS 2020,
  December 6-12, 2020, virtual}, 2020.

\bibitem[Sohn et~al.(2020)Sohn, Berthelot, Carlini, Zhang, Zhang, Raffel,
  Cubuk, Kurakin, and Li]{fixmatch}
Sohn, K., Berthelot, D., Carlini, N., Zhang, Z., Zhang, H., Raffel, C., Cubuk,
  E.~D., Kurakin, A., and Li, C.
\newblock Fixmatch: Simplifying semi-supervised learning with consistency and
  confidence.
\newblock In Larochelle, H., Ranzato, M., Hadsell, R., Balcan, M., and Lin, H.
  (eds.), \emph{Advances in Neural Information Processing Systems 33: Annual
  Conference on Neural Information Processing Systems 2020, NeurIPS 2020,
  December 6-12, 2020, virtual}, 2020.

\bibitem[Tarvainen \& Valpola(2017)Tarvainen and Valpola]{mean_teacher}
Tarvainen, A. and Valpola, H.
\newblock Mean teachers are better role models: Weight-averaged consistency
  targets improve semi-supervised deep learning results.
\newblock In Guyon, I., von Luxburg, U., Bengio, S., Wallach, H.~M., Fergus,
  R., Vishwanathan, S. V.~N., and Garnett, R. (eds.), \emph{Advances in Neural
  Information Processing Systems 30: Annual Conference on Neural Information
  Processing Systems 2017, December 4-9, 2017, Long Beach, CA, {USA}}, pp.\
  1195--1204, 2017.

\bibitem[Tran et~al.(2019)Tran, Tran, Nguyen, Yang, and Cheung]{ssgan_minimax}
Tran, N., Tran, V., Nguyen, N., Yang, L., and Cheung, N.
\newblock Self-supervised {GAN:} analysis and improvement with multi-class
  minimax game.
\newblock In Wallach, H.~M., Larochelle, H., Beygelzimer, A.,
  d'Alch{\'{e}}{-}Buc, F., Fox, E.~B., and Garnett, R. (eds.), \emph{Advances
  in Neural Information Processing Systems 32: Annual Conference on Neural
  Information Processing Systems 2019, NeurIPS 2019, December 8-14, 2019,
  Vancouver, BC, Canada}, pp.\  13232--13243, 2019.

\bibitem[van Engelen \& Hoos(2020)van Engelen and Hoos]{SSLsurvey2020}
van Engelen, J.~E. and Hoos, H.~H.
\newblock A survey on semi-supervised learning.
\newblock \emph{Mach. Learn.}, 109\penalty0 (2):\penalty0 373--440, 2020.

\bibitem[Zhao et~al.(2020{\natexlab{a}})Zhao, Liu, Lin, Zhu, and
  Han]{diffaug_gan}
Zhao, S., Liu, Z., Lin, J., Zhu, J., and Han, S.
\newblock Differentiable augmentation for data-efficient {GAN} training.
\newblock In Larochelle, H., Ranzato, M., Hadsell, R., Balcan, M., and Lin, H.
  (eds.), \emph{Advances in Neural Information Processing Systems 33: Annual
  Conference on Neural Information Processing Systems 2020, NeurIPS 2020,
  December 6-12, 2020, virtual}, 2020{\natexlab{a}}.

\bibitem[Zhao et~al.(2020{\natexlab{b}})Zhao, Li, Yu, Gao, and Chen]{fq_gan}
Zhao, Y., Li, C., Yu, P., Gao, J., and Chen, C.
\newblock Feature quantization improves {GAN} training.
\newblock In \emph{Proceedings of the 37th International Conference on Machine
  Learning, {ICML} 2020, 13-18 July 2020, Virtual Event}, volume 119 of
  \emph{Proceedings of Machine Learning Research}, pp.\  11376--11386. {PMLR},
  2020{\natexlab{b}}.

\bibitem[Zhao et~al.(2020{\natexlab{c}})Zhao, Singh, Lee, Zhang, Odena, and
  Zhang]{icr_gan}
Zhao, Z., Singh, S., Lee, H., Zhang, Z., Odena, A., and Zhang, H.
\newblock Improved consistency regularization for {GANs}.
\newblock \emph{CoRR}, abs/2002.04724, 2020{\natexlab{c}}.

\end{thebibliography}
\bibliographystyle{icml2021}

\clearpage

\end{document}